\documentclass{article}

\PassOptionsToPackage{numbers, sort&compress}{natbib}

\usepackage[final]{neurips_2025_ml4ps}

\usepackage[utf8]{inputenc} 
\usepackage[T1]{fontenc}    
\usepackage{hyperref}       
\usepackage{url}            
\usepackage{booktabs}       
\usepackage{amsfonts}       
\usepackage{nicefrac}       
\usepackage{microtype}      
\usepackage{xcolor}         

\newcommand{\bestMNISTFullAccuracy}{97.2±0.1\%}
\newcommand{\noiselessBestMNISTHundredAccuracy}{90.6±1.7\%}
\newcommand{\noisyBestMNISTHundredAccuracy}{92.0±0.3\%}
\newcommand{\bestFashionMNISTFullAccuracy}{88.0±0.1\%}

\newcommand{\bestBPTTMNISTFullAccuracy}{96.8±0.1\%}

\newcommand{\FourBitPhaseQuantisationAccuracy}{89.8±1.5\%}

\newcommand{\TenBitParameterQuantisationMNISTHundredAccuracy}{89.4±1.5\%}

\usepackage{amsmath}
\usepackage{amssymb}
\usepackage{graphicx}

\title{Learning at the Speed of Physics: Equilibrium Propagation on Oscillator Ising Machines}

\author{Alex Gower, \\
University of Cambridge \& Nokia Bell Labs,  \\
apg59@cam.ac.uk, \\
alexanderpgower@gmail.com}

\begin{document}

\maketitle

\begin{abstract}
Physical systems that naturally perform energy descent offer a direct route to accelerating machine learning. Oscillator Ising Machines (OIMs) exemplify this idea: their GHz-frequency dynamics mirror both the optimization of energy-based models (EBMs) and gradient descent on loss landscapes, while intrinsic noise corresponds to Langevin dynamics—supporting sampling as well as optimization. Equilibrium Propagation (EP) unifies these processes into descent on a single total energy landscape, enabling local learning rules without global backpropagation. We show that EP on OIMs achieves competitive accuracy ($\sim$\bestMNISTFullAccuracy{} on MNIST, $\sim$\bestFashionMNISTFullAccuracy{} on Fashion-MNIST), while maintaining robustness under realistic hardware constraints such as parameter quantization and phase noise. These results establish OIMs as a fast, energy-efficient substrate for neuromorphic learning, and suggest that EBMs—often bottlenecked by conventional processors—may find practical realization on physical hardware whose dynamics directly perform their optimization.
\end{abstract}

\section{Introduction}

Equilibrium Propagation (EP) is a local alternative to backpropagation, avoiding separate backward passes by nudging steady states toward desired outputs. It is particularly suited to physical systems that naturally perform energy descent toward local minima.

Oscillator Ising Machines (OIMs) are natural candidates for EP because their dynamics follow energy gradient descent~\cite{basharNoteAnalyzingStability2023}. Unlike discrete Ising solvers~\cite{laydevantTrainingIsingMachine2024}, OIMs have continuous phase dynamics that support the precise nudging required by EP, with couplings and biases acting as trainable parameters. Originally developed to minimize Ising energies, they can now be repurposed as neuromorphic machine learning processors.

The synergy between EP and OIMs offers the prospect of fast, energy-efficient neuromorphic computing. GPU training consumes substantial power~\cite{garcia-martinEstimationEnergyConsumption2019}, while OIMs achieve both speed and energy efficiency by exploiting GHz-frequency physical dynamics for rapid energy descent~\cite{csabaCoupledOscillatorsComputing2020b}. This acceleration could make energy-based learning practical, overcoming the long relaxation and sampling times that limit such models on conventional processors~\cite{assranSelfSupervisedLearningImages2023}.

Previous EP implementations on oscillators faced initialization or synchronization issues~\cite{wangTrainingCoupledPhase2024,rageauTrainingSynchronizingOscillator2025}. We present the first framework that maps EP directly onto existing OIM hardware without modification, turning devices originally designed for combinatorial optimization into neural network processors; a more detailed write-up of this study is available in~\cite{gowerOIMEPFull}. This illustrates a broader principle: when physical dynamics coincide with ML optimization, they can be directly harnessed to build faster and more efficient learning systems.

\section{Oscillator Ising Machines}

An Oscillator Ising Machine (OIM) consists of a network of $n$ coupled self-sustaining nonlinear oscillators with non-varying amplitudes and equal frequencies. Over a sufficiently coarse-grained timescale~\cite{main-OIMs-paper}, each oscillator can be parameterized by a phase $\phi_i \in [0,2\pi]$ whose dynamics follow $\frac{d\phi_i}{dt'} = -\sum_{j=1, \ j\neq i}^n J_{ij} \sin(\phi_i - \phi_j ) -h_i \sin(\phi_i) - S_i \sin(2 \phi_i)$.

Here, $J_{ij}$ is the coupling strength, $h_i$ a bias favoring phase alignment with either $0$ or $\pi$, $S_i$ a synchronization field encouraging binary ($0/\pi$) phase states, and $t'$ a dimensionless time such that real time $t = t'/\bar\omega$ scales inversely with oscillator frequencies.

These dynamics are equivalent to gradient descent $\frac{d \phi_i}{dt'} = - \frac{\partial V}{\partial \phi_i}$, for an energy function\footnote{Symmetric $J_{ij}$ includes $\frac{1}{2}$ to avoid double counting.}~\cite{basharNoteAnalyzingStability2023}: $V = -\frac{1}{2}\sum_{i,j \ i \neq j}^n J_{ij} \cos(\phi_i-\phi_j) - \sum_i^n h_i \cos(\phi_i) - \sum_i^n \frac{S_i}{2} \cos(2\phi_i).$

This rare energy-descent property makes OIMs well suited as neuromorphic ML processors.

\section{Equilibrium Propagation}\label{sec:equilibrium-propagation}

Equilibrium Propagation (EP)~\cite{scellierEquilibriumPropagationBridging2017,laborieuxScalingEquilibriumPropagation2020} is a local alternative to backpropagation for networks whose dynamics are characterized by the convergence to a stationary point, avoiding a separate backward pass by nudging these steady states toward desired outputs.

For systems with energy gradient descent dynamics, EP utilizes a `total energy' $F(x,s, \{\theta\})$, which depends on inputs $x$, trainable parameters $\{\theta\}$, and a dynamical state $s$ that includes both hidden and output variables ($s=(h,y)$). The system follows gradient descent dynamics: $\frac{ds}{dt} = - \frac{\partial F}{\partial s}$ (for fixed $x$ and $\{ \theta \}$).

This total energy decomposes as $F = E + \beta \ell$, where $E(x,s,\{\theta\})$ is the free system's energy function, $\ell(y,\hat y)$ is a loss function comparing outputs $y$ to targets $\hat y$, and $\beta$ is the nudging factor that biases dynamics toward configurations that minimize loss.

For each training example $x$, EP proceeds in three phases: 
(i) Free phase: Initialize at a reference state $s_0$ with $\beta=0$ and evolve to stationary point $s_*$; (ii) Positive nudged phase: From $s_*$, evolve with $\beta > 0$ to reach stationary point $s_*^{\beta}$; (iii) Negative nudged phase: From $s_*$ again, evolve with $\beta < 0$ to reach stationary point $s_*^{-\beta}$.


The updates to the trainable parameters $\{ \theta \}$ are then given by~\cite{laborieuxScalingEquilibriumPropagation2020} $\Delta \theta = \eta \hat{\nabla}^{\rm EP}(\beta)$ for learning rate $\eta$ and where:
\begin{equation}\label{eq:ep-parameter-update}
    \hat{\nabla}^{\rm EP}(\beta) = -\frac{1}{2\beta} \left( \frac{\partial F}{\partial \theta}(x, s^{\beta}_*, \{\theta\}) - \frac{\partial F}{\partial \theta}(x, s_*^{-\beta}, \{\theta\}) \right).
\end{equation}

This update effectively steers the free stationary point toward configurations with lower loss by decreasing energy at $s^\beta_*$ and increasing it at $s^{-\beta}_*$. In the limit of infinitesimal $\beta$, this update equals gradient descent on the loss~\cite{laborieuxScalingEquilibriumPropagation2020}:
\begin{equation}
   \lim_{\beta \to 0} \hat{\nabla}^{\rm EP}(\beta) = - \frac{\partial \ell}{\partial \theta}(y_*, \hat y).
\end{equation}

Moreover, during the nudged phase dynamics (before convergence), the instantaneous EP updates match Backpropagation Through Time (BPTT) in the $\beta \to 0$ limit~\cite{werbosBackpropagationTimeWhat1990}:
\begin{equation}
 \lim_{\beta \to 0} -\frac{1}{2\beta} \left[ 
 \frac{\partial F}{\partial \theta}(x, s_t^{\beta}, \{\theta\}) -
 \frac{\partial F}{\partial \theta}(x, s_t^{-\beta}, \{\theta\})
 \right] = \hat{\nabla}^{\rm BPTT}(t).
\end{equation}
We verify this EP-BPTT correspondence empirically in Section~\ref{sec:results}.

\section{Implementation on Oscillator Ising Machines}\label{sec:implementation-on-oims}

We implement a dense MLP neural network on an OIM with $n_x$ input neurons (fixed during dynamics), $n_h$ hidden neurons, and $n_y$ output neurons. Each non-input neuron corresponds to an oscillator, with the dynamical state $\phi=(\phi^{(h)},\phi^{(y)})$, and trainable parameters ${\theta}=({w_{ij}^{(x,h)}},{w_{ij}^{(h,y)}},{b_i^{(h)}},{b_i^{(y)}})$.

We implement neural network components through specific energy terms in the OIM: biases via $-b_i \cos(\phi_i)$ terms, and hidden-to-output weights via $-w^{(h,y)}_{ij} \cos(\phi_i^{(h)} - \phi_j^{(y)})$ terms. Using the mapping $s_i = \cos(\phi_i)$ to transform phases into values in $[-1,1]$, we implement input-to-hidden weights through $-w^{(x,h)}_{ij} x_i \cos(\phi^{(h)}_j)$ terms.

For an MSE loss function $\ell(y,\hat y) = \frac{1}{2} \sum_i(y_i-\hat y_i)^2$ with $y_i=\cos(\phi^{(y)}_i)$, each term expands as $\frac{1}{2}\big( \cos(\phi^{(y)}_i) - \hat y_i \big)^2 = \frac{1}{4}\cos(2\phi_i^{(y)}) - \hat y_i\cos(\phi^{(y)}_i) + \text{const.}$ (using $\cos^2(x) = \frac{1}{2}[1 + \cos(2x)]$).

These terms align with the OIM energy $V$; thus the total energy $F=E+\beta\ell$ is implemented by setting: $ h_i^{(h)} = b_i^{(h)}+\sum_{j=1}^{n_x}w_{ji}^{(x,h)}x_j, \quad J_{ij}^{(h,y)} = w_{ij}^{(h,y)}, \quad h_i^{(y)} = b_i^{(y)}+\beta \hat y_i, \quad
S_i^{(y)} = -\tfrac{\beta}{2}.$

Note that input-to-hidden weights enter via hidden-bias terms $h_i^{(h)}$, so no extra oscillators are needed. Synchronization fields $S_i^{(y)}$, conveniently already present in OIMs, are needed to implement the MSE loss at outputs.

Training follows standard EP. We use He weight initialization\cite{heDelvingDeepRectifiers2015} with zero biases. For each input $x$: (i) Configure the OIM parameters as specified above; (ii) Initialize all oscillators to the reference state\footnote{This corresponds to setting all neuron activations $\cos(\frac{\pi}{2})=0$. We found that any random initialization of phases also works well, provided the same random reference state is used consistently across all dynamics.} $\phi_0 = \{\frac{\pi}{2}\}$; (iii) Run the free phase ($\beta=0$) dynamics until convergence to $\phi_*$; (iv) Run the positive and negative nudged phase dynamics (initialized at $\phi_*$) until convergence to $\phi_*^{+\beta}$ and $\phi_*^{-\beta}$.

The EP update rule gives the following local parameter updates:
\begin{align*}
\Delta b^{(h)}_{i} &\propto -\tfrac{1}{2\beta}\!\left[\cos(\phi_i^{(h),-\beta}) - \cos(\phi_i^{(h),+\beta})\right], 
\ \ \Delta b^{(y)}_{i} \propto -\tfrac{1}{2\beta}\!\left[\cos(\phi_i^{(y),-\beta}) - \cos(\phi_i^{(y),+\beta})\right], \\
\Delta w_{ij}^{(x,h)} &\propto -\tfrac{1}{2\beta}\!\left[x_i\cos(\phi^{(h),-\beta}_j) - x_i\cos(\phi^{(h),+\beta}_j)\right], \\
 \Delta w^{(h,y)}_{ij} &\propto -\tfrac{1}{2\beta}\!\left[\cos(\phi^{(h),-\beta}_i-\phi^{(y),-\beta}_j) - \cos(\phi^{(h),+\beta}_i-\phi^{(y),+\beta}_j)\right].
\end{align*}

All updates are local, depending only on phases of connected oscillators, measurable at each synapse or neuron. This locality eliminates global backpropagation circuitry. Updates are averaged over mini-batches and remain compatible with standard optimizers (learning-rate schedules, weight decay, momentum).

\section{Results}\label{sec:results}

We validate our OIM–EP implementation on MNIST~\cite{lecunGradientbasedLearningApplied1998}, Fashion-MNIST~\cite{xiaoFashionMNISTNovelImage2017}, and the MNIST/100 subset (1,000/100 train/test used in prior Ising machine benchmarks~\cite{laydevantTrainingIsingMachine2024}).

\begin{figure}[!t]
  \centering
  \begin{minipage}{0.49\linewidth}
    \centering
    \includegraphics[width=\linewidth]{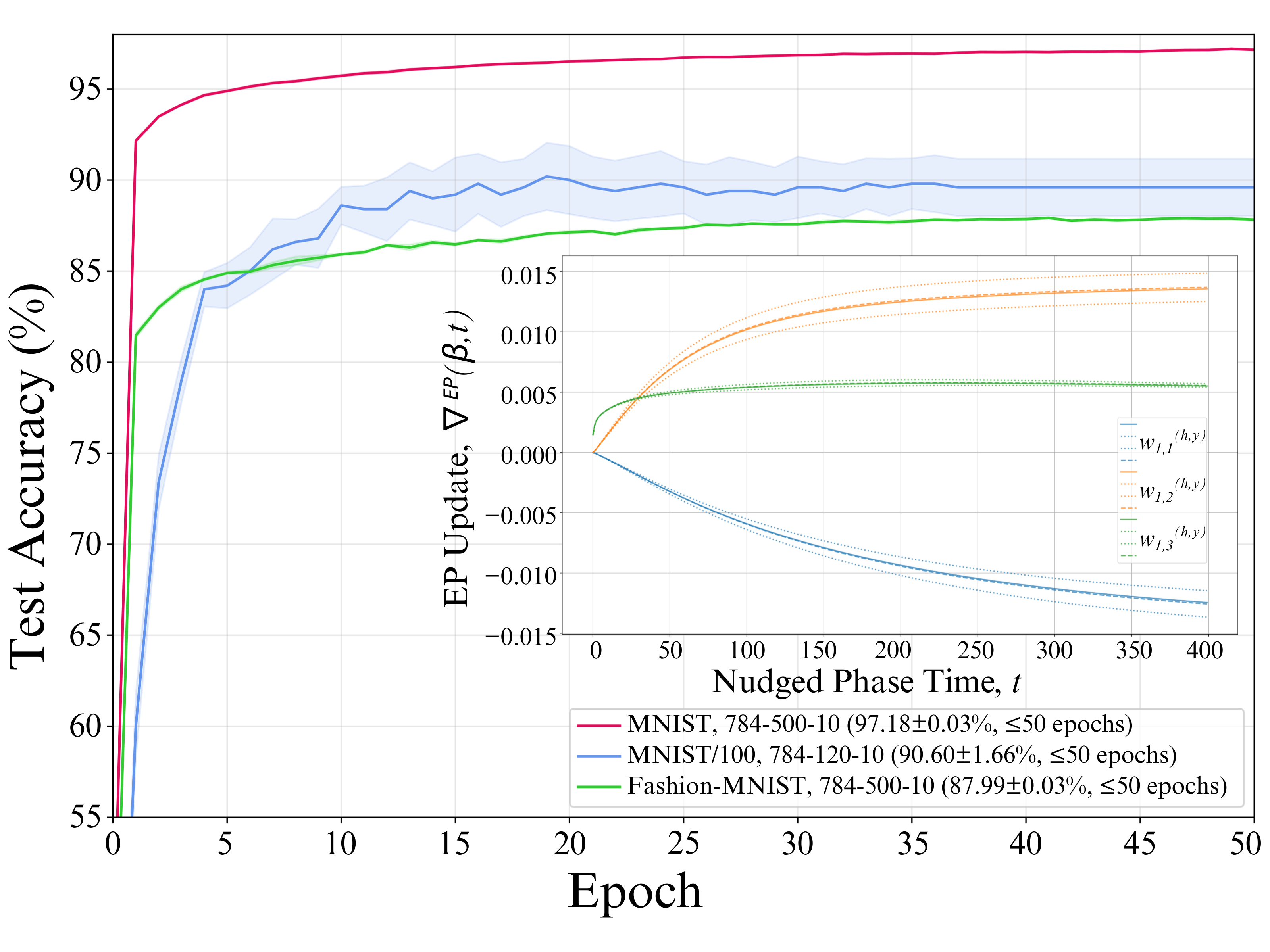}
  \end{minipage}\hfill
  \begin{minipage}{0.49\linewidth}
    \centering
    \includegraphics[width=\linewidth]{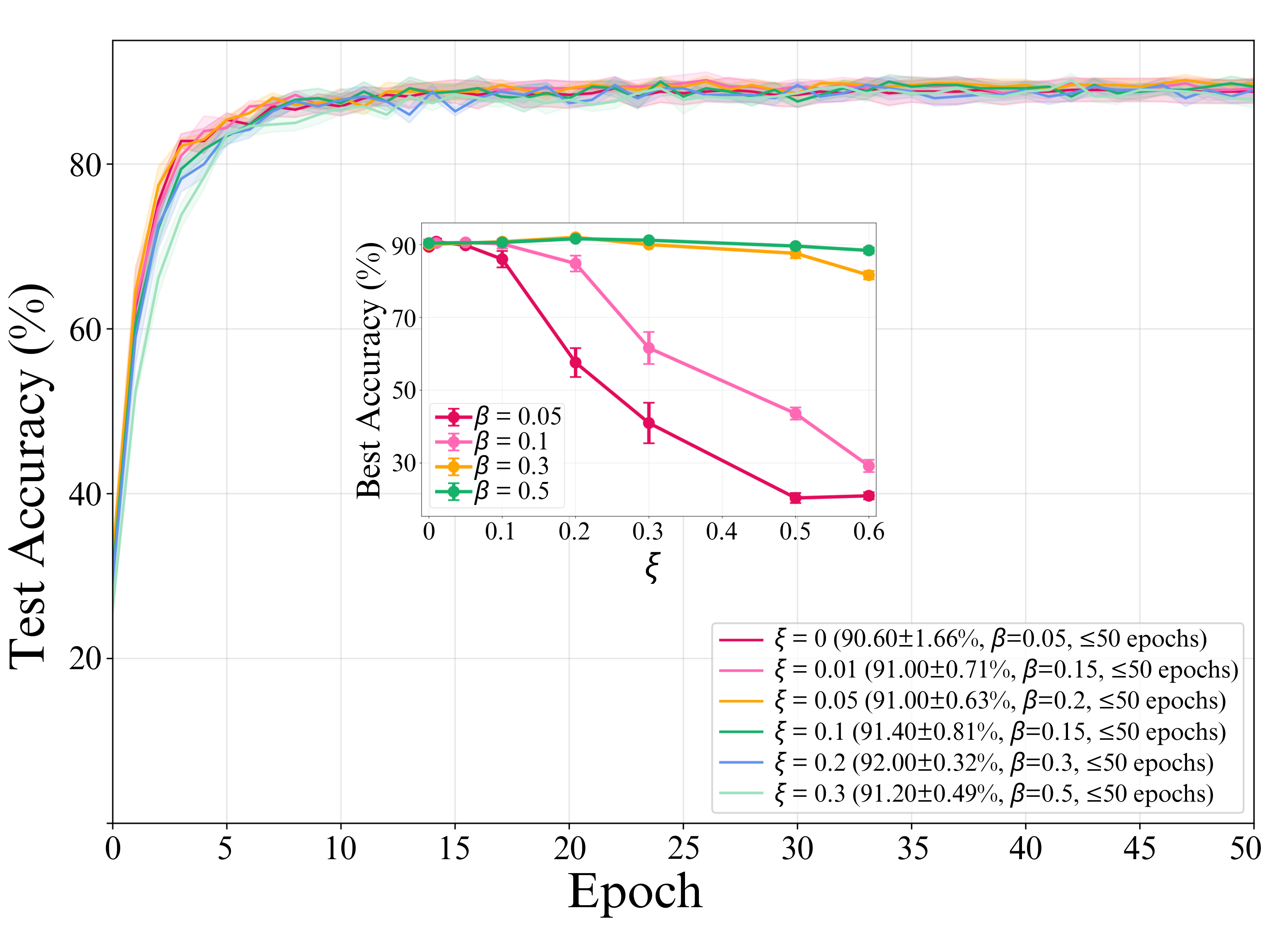}
  \end{minipage}
  \vspace{-2mm}
  \caption{(Left) Accuracy: 784–500–10 reaches \bestMNISTFullAccuracy{} (MNIST) and \bestFashionMNISTFullAccuracy{} (FMNIST); 784–120–10 reaches \noiselessBestMNISTHundredAccuracy{} (MNIST/100). Inset: EP--BPTT correspondence, with symmetric EP (dashed) matching BPTT (solid). (Right) Noise robustness (784–120–10): accuracy remains high with optimized $\beta$ (peak \noisyBestMNISTHundredAccuracy{} at $\xi=0.2$). Inset: fixed-$\beta$ training degrades at large noise $\xi$.}
  \label{fig:training-noise}
  \vspace{-3mm}
\end{figure}

Training used PyTorch\footnote{Code available at
\url{https://github.com/alexgower/OIM-Equilibrium-Propagation}.} with explicit Euler
integration of the OIM dynamics. For full MNIST/FMNIST we used
$T=4000$ free-phase steps and $K=400$ nudged steps with time step
$\epsilon=0.45$; for MNIST/100 we used $T=3500$, $K=350$,
$\epsilon=0.5$. We used constant nudging factors $\beta=0.1$
(full MNIST/FMNIST) and $\beta=0.05$ (MNIST/100) unless otherwise
stated. The parameters
$(T,K,\epsilon,\beta)$ were selected by increasing the free-phase
duration $\epsilon T$ and decreasing $\beta$ until the symmetric EP
updates closely matched BPTT updates (EP--BPTT inset of
Fig.~\ref{fig:training-noise}) while maintaining stable dynamics, and by
choosing the nudged-phase duration $\epsilon K$ large enough that the
nudged updates saturated. We used fixed-length trajectories and verified that $\max_i |d\phi_i/dt'|$ was negligible
at the end of each free phase, indicating convergence. We used a batch
size of 128 for full MNIST/FMNIST and 20 for MNIST/100, and
layer-specific learning rates
$(\eta^{(h)}_{w},\eta^{(y)}_{w},\eta^{(h)}_{b},\eta^{(y)}_{b})
= (0.01, 0.001, 0.001, 0.001).$

A key validation is that EP parameter updates match Backpropagation Through Time (BPTT). Fig.~\ref{fig:training-noise} shows this correspondence (inset), confirming OIMs satisfy the gradient-descent property needed for scaling, despite their nonlinear sinusoidal couplings which differ from the quadratic couplings used in conventional EP implementations.

\begin{figure}[!t]
  \centering
  \begin{minipage}{0.49\linewidth}
    \centering
    \includegraphics[width=\linewidth]{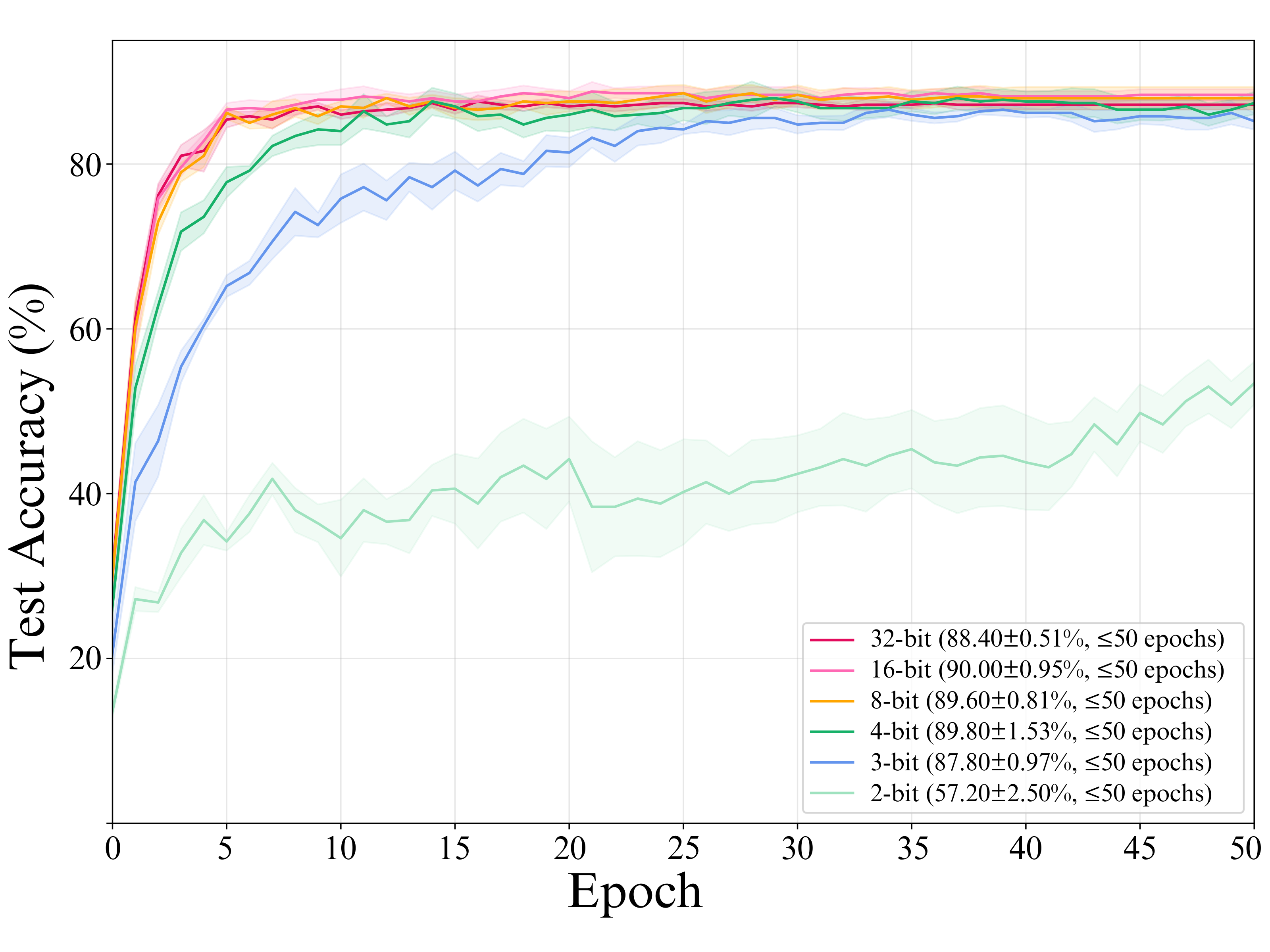}
  \end{minipage}\hfill
  \begin{minipage}{0.49\linewidth}
    \centering
    \includegraphics[width=\linewidth]{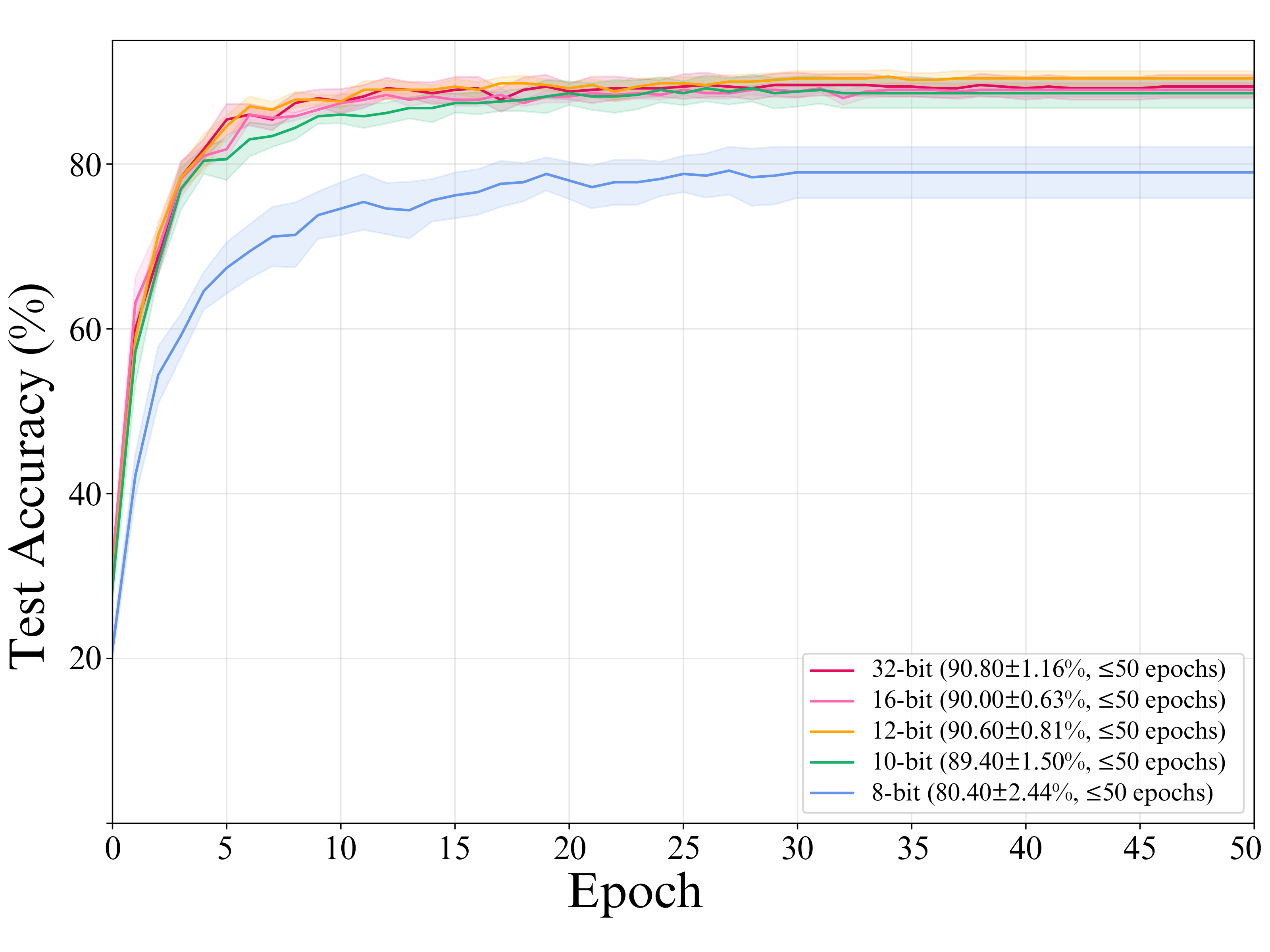}
  \end{minipage}
  \vspace{-2mm}
  \caption{(Left) Phase quantization (784–120–10): accuracy holds down to 4-bit (\FourBitPhaseQuantisationAccuracy), degrades at 2-bit. (Right) Parameter quantization (784–120–10): stable at 10-bit (\TenBitParameterQuantisationMNISTHundredAccuracy), degrades at 8-bit.}
  \label{fig:quantization}
  \vspace{-3mm}
\end{figure}

OIM–EP MLPs with a 784–120–10 architecture achieve
\noiselessBestMNISTHundredAccuracy{} on MNIST/100
(\noisyBestMNISTHundredAccuracy{} with noise), and a 784–500–10 MLP
reaches \bestMNISTFullAccuracy{}/\bestFashionMNISTFullAccuracy{} on full
MNIST/FMNIST (Fig.~\ref{fig:training-noise}). These results surpass D-Wave annealers~\cite{laydevantTrainingIsingMachine2024} and match BPTT on the same OIMs (\bestBPTTMNISTFullAccuracy{} on MNIST), indicating accuracy is architecture-limited rather than training-limited. On FMNIST, OIMs outperform p-bit Ising machines~\cite{niaziTrainingDeepBoltzmann2024} and are competitive with other neuromorphic MLPs~\cite{zhangTemporalSpikeSequence2020}.

Hardware viability requires robustness. With Gaussian phase noise (Fig.~\ref{fig:training-noise}), accuracy degrades if $\beta$ is fixed, but remains robust when $\beta$ is optimized, peaking at \noisyBestMNISTHundredAccuracy{} for $\xi=0.2$. Robustness holds for $\beta \gtrsim \xi/2$, suggesting moderate noise can act as useful regularization. With phase quantization, performance holds down to 4-bit precision (Fig.~\ref{fig:quantization}); with parameter quantization it holds at 10-bit. 

Overall, OIMs maintain accuracy under 10-bit parameters, 4-bit phase detection~\cite{graberIntegratedCoupledOscillator2024}, and moderate phase noise, confirming viability for CMOS implementations~\cite{royExperimentsOscillatorBased2025}.

\section{Discussion}

Our results show that OIMs can implement Equilibrium Propagation with accuracy comparable to conventional methods while offering potential advantages in speed, energy efficiency, and scalability. Robustness under realistic hardware limits—10-bit parameters, 4-bit phase detection, and tolerance to moderate noise—indicates viability for CMOS implementations.

Physical OIMs promise dramatic speedups. With GHz-frequency oscillators reaching microsecond timescales for parameter updates and phase measurements~\cite{main-OIMs-paper}, full MNIST training (50 epochs, 60k examples, 3 EP phases) could complete in seconds to minutes, compared to $\sim$40 hours for EP and $\sim$60 hours for BPTT in our simulations—an acceleration of several orders of magnitude. This speed and efficiency could make energy-based learning practical for tasks previously limited by training time.

Noise tolerance further suggests that hardware need not eliminate fluctuations entirely. With $\beta \gtrsim \xi/2$, accuracy is preserved even at $\xi=0.3$, reducing design overhead and potentially exploiting noise as a form of regularization. Notably, OIM noise also corresponds to Langevin dynamics, hinting at native support for sampling as well as optimization.

Finally, EP requires no major hardware modifications: OIMs already provide coupling, biases, synchronization, and calibrated oscillators. Devices built for combinatorial optimization can thus be repurposed as neuromorphic processors through software-level configuration alone—illustrating the broader principle that physical dynamics aligned with ML objectives can directly accelerate learning and inference.

\section*{Acknowledgements}

This work was supported by UKRI/EPSRC CASE Award 220191 in partnership
with Nokia UK Limited, and by a G-Research academic grant, which helped
support the presentation of this work at NeurIPS 2025.

\bibliographystyle{abbrvnat}
\bibliography{references}

\end{document}